\documentclass[10pt,twocolumn,letterpaper]{article}

\usepackage{wacv}              

\usepackage{graphicx}
\usepackage{amsmath}
\usepackage{amssymb}
\usepackage{booktabs}
\usepackage{times}
\usepackage{epsfig}
\usepackage{setspace} 
\usepackage{zref-xr}
\usepackage{tablefootnote}
\usepackage{xcolor}
\usepackage{enumerate}
\usepackage{float}
\usepackage{chngcntr}
\usepackage{soul}
\usepackage{caption}
\usepackage{subcaption}
\usepackage[accsupp]{axessibility} 

\usepackage[pagebackref,breaklinks,colorlinks]{hyperref}

\usepackage[capitalize]{cleveref}
\crefname{section}{Sec.}{Secs.}
\Crefname{section}{Section}{Sections}
\Crefname{table}{Table}{Tables}
\crefname{table}{Tab.}{Tabs.}

\usepackage{pifont}
\newcommand{\cmark}{\ding{51}}%
\newcommand{\xmark}{\ding{55}}%

\def\wacvPaperID{1281} 
\def\confName{WACV}
\def\confYear{2024}

\begin{document}

\title{Object-centric Video Representation for Long-term Action Anticipation}

\author{
Ce Zhang$^{1}$\thanks{The first two authors contribute equally.}
\qquad
Changcheng Fu$^{1*}$
\qquad
Shijie Wang$^{1}$
\qquad
Nakul Agarwal$^{2}$\\
Kwonjoon Lee$^{2}$
\qquad
Chiho Choi$^{3}$\thanks{Work done while at Honda Research Institute USA.}
\qquad
Chen Sun$^{1}$
 \\
$^{1}$ Brown University \quad $^{2}$ Honda Research Institute USA \quad $^{3}$ Samsung
}
\maketitle

\begin{abstract}

This paper focuses on building object-centric representations for long-term action anticipation in videos. Our key motivation is that objects provide important cues to recognize and predict human-object interactions, especially when the predictions are longer term, as an observed ``background'' object could be used by the human actor in the future.
We observe that existing object-based video recognition frameworks either assume the existence of in-domain supervised object detectors or follow a fully weakly-supervised pipeline to infer object locations from action labels. We propose to build object-centric video representations by leveraging visual-language pretrained models. This is achieved by ``object prompts'', an approach to extract task-specific object-centric representations from general-purpose pretrained models without finetuning. To recognize and predict human-object interactions, we use a Transformer-based neural architecture which allows the ``retrieval'' of relevant objects for action anticipation at various time scales. We conduct extensive evaluations on the Ego4D, 50Salads, and EGTEA Gaze+ benchmarks. Both quantitative and qualitative results confirm the effectiveness of our proposed method. Our code is available at \href{https://github.com/brown-palm/ObjectPrompt}{github.com/brown-palm/ObjectPrompt}.

\end{abstract}
\section{Introduction}
\label{sec:intro}

\begin{figure}[h]
  \centering
   \includegraphics[width=1.05\linewidth]{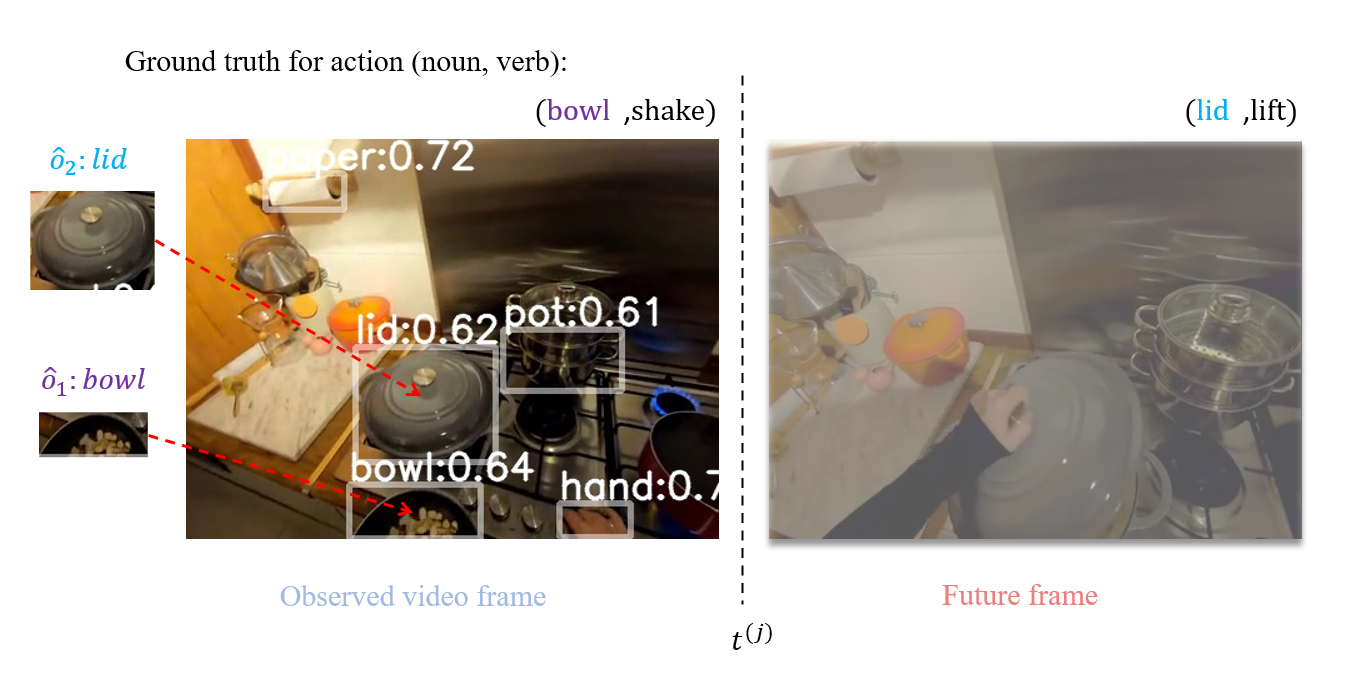}
\vspace{-2em}
   \caption{Objects are not only helpful for action recognition (left, \textit{shake a bowl}), they also reveal the possible options for future human-object interactions (right, \textit{lift a lid}). We propose object prompts, which leverage visual-language pretrained models (\eg GLIP~\cite{glip}) to build object-centric video representations without dataset-specific finetuning.}
\vspace{-2em}
   \label{fig:object_visualization}
\end{figure}

Given an egocentric video observation, the action anticipation task~\cite{salad_dataset2013} is defined as generating an action sequence of the camera-wearer in the form of verb and object pairs. Of particular interest is the long-term action anticipation (LTA) task~\cite{ego4d}, which aims to anticipate future actions over a long time-horizon. A reliable action anticipation algorithm is crucial for building intelligent agents, as it provides important signals for planning in interactive environments.  

This paper aims to build effective object-centric video representations for action anticipation. As illustrated in Figure~\ref{fig:object_visualization}, our key motivation is that a detailed, object-centric understanding of the scene provides visual cues on the goals of the actions and the available tools to be interacted with. The object-centric representation is also inspired and supported by studies on human perception~\cite{kahneman1992reviewing,grill2005visual,tenenbaum2011grow}.

While objects have been shown to play a crucial role for action understanding in both humans~\cite{10.3389/fpsyg.2016.00111,Woodward1998-WOOISE} and machines~\cite{zhang2022object,materzynska2020something}, their impacts on video-based action anticipation are yet to be studied.
Among the earlier attempts at object-based video action understanding, one common approach is to leverage object detectors trained on in-domain bounding box annotations~\cite{Epic-Kitchen,herzig2022object,zhang2022object,materzynska2020something,zhou2023can}  on the same or similar datasets. While effective, the in-domain bounding box annotation process is time-consuming and labor-intensive, which makes the corresponding frameworks unlikely to scale to visually more diverse videos and complex, cluttered scenes. The other approach is to leverage generic object proposals~\cite{wang2018videos,sun2019relational}, or to directly work with image patches~\cite{sun2018actor,patrick2021keeping,girdhar2019video}, and rely on the attention mechanism~\cite{transformer} to pick the salient regions with weak supervision from the action labels. Despite being more flexible, this approach does not incorporate prior knowledge on object locations, and often struggles to ``detect'' the actual objects, especially with limited training data.

We propose to address the limitations of the existing approaches by leveraging visual-language models pretrained on large-scale datasets, such as GLIP~\cite{glip}. We hypothesize that these pretrained models, whose objective is to (contrastively) associate image regions with text descriptions, learn generic object-centric representation that can be transferred to the action anticipation task, without the need to finetune their weights. We identify two challenges in order to investigate whether visual-language pretrained models help anticipation: First, how to properly ``query'' the pretrained models to retrieve the most relevant objects in a video observation, based on the domain knowledge; Second, how to associate different objects in an often cluttered scene to predict different human-object interactions in a long time-horizon. For the first challenge, we propose \textit{object prompts}, which incorporate the domain knowledge of the target dataset by mapping the action (\eg verb and object pairs) vocabulary into object prompts. The prompts are used to query the visual-language pretrained models to focus on the objects of interest. For the second challenge, we propose using a \textit{predictive transformer encoder} (PTE), which is a Transformer encoder network that jointly attends to the motion cues (based on pre-trained video encoders) and the object-centric representation (based on the object prompts), and dynamically associates the motion and object evidence in order to predict future actions at different time steps. The proposed framework is trained end-to-end with the future action classification objectives. We refer to our overall framework as \textbf{ObjectPrompt}.

We conduct thorough experiments on the Ego4D~\cite{ego4d} and the 50Salads~\cite{salad_dataset2013} long-term action anticipation benchmarks, and the anticipation benchmark EGTEA Gaze+~\cite{li2018eye}. Our quantitative experiments confirm that the prompt-based object-centric representations can substantially improve the action anticipation performance. Ablation experiments reveal that it is important to incorporate in-domain knowledge when designing the object prompts, and that object-centric representation significantly outperforms image-level counterpart~\cite{das2022video+}. In addition, qualitative analysis shows that the model learns to associate the corresponding objects when predicting actions at different time steps. 

In summary, our contributions in this paper are threefold: First, we demonstrate the effectiveness of object-centric representation for video action anticipation; Second, we propose ``object prompts'' to incorporate domain information when querying the pretrained models, and predictive transformer encoder to dynamically associate the object evidence for action anticipation; Finally, we provide extensive quantitative and qualitative analysis of the proposed framework, which achieves competitive performance on three benchmarks. Our implementation along with the pretrained models will be released.

\section{Related Work}
\label{sec:related}

\noindent\textbf{Object-centric Video Representation} is an active research area for action recognition applications. The motivations include producing a more compact, thus efficient video representation by attending to regions of interest; and enabling compositional generalization to unseen human-object interactions~\cite{CVPR2020_SomethingElse} with a structured representation.

For example, Wang et al.~\cite{wang2018videos} extract RoI features from an 3D CNN feature maps using off-the-shelf detector, and build a graph neural network on top of RoI features alone.
Long-term Feature Banks (LFB)~\cite{wu2019long} and Object Transformer~\cite{wu2021towards} load off-the-shelf object features from object detectors in video backbones to encode long-term video features.
Recently, with the advance of transformer-based architectures,
ORViT~\cite{herzig2022object} proposes to crop~\cite{he2017mask} object regions as a new object tokens and attach them to pixel tokens.
ObjectLearner~\cite{zhang2022object} fuses an object-layout stream and a pixels stream using an object-to-pixel transformer.
Most of the existing object-based approaches either assume the availability of in-domain bounding box annotations, or apply generic ``objectness'' criteria (\eg bounding box proposals trained on COCO~\cite{COCO}). ObjectViViT~\cite{zhou2023can} relies on object detectors trained on the target datasets, or ground truth object annotations during evaluation. They observe that a direct application of a general-purpose large-vocabulary object detector~\cite{zhou2022detecting} does not help video classification. We propose object prompts to leverage a pretrained visual-language model to generate task-specific object representations, which offers a simple yet effective solution to incorporate domain knowledge when they are available.

\noindent\textbf{Video Transformers.} Transformers~\cite{dosovitskiy2020image} are now the predominant architectures for video recognition.
The vanilla vision transformer (ViT)~\cite{dosovitskiy2020image} evenly devides images into non-overlapping tokens, and run multi-head self-attention~\cite{transformer} over the tokens. 
TimeSFormer~\cite{bertasius2021space} and ViViT~\cite{arnab2021vivit} extends ViT to videos, by introducing cube tokenization and efficient cross-time attention, i.e., axis-based space-time attention~\cite{bertasius2021space} or factorized attention~\cite{arnab2021vivit}.
MotionFormer~\cite{patrick2021keeping} enhance space-time attention by an implicit trajectory attention module.
MViT~\cite{fan2021multiscale,li2021improved} and VideoSwin Transformer~\cite{liu2021video} re-introduce resolution-pooling as in convolution nets in video transformers for efficiency.

\begin{figure*}[ht]
  \centering
  \begin{subfigure}{.9\linewidth}
    \includegraphics[width=1\linewidth]{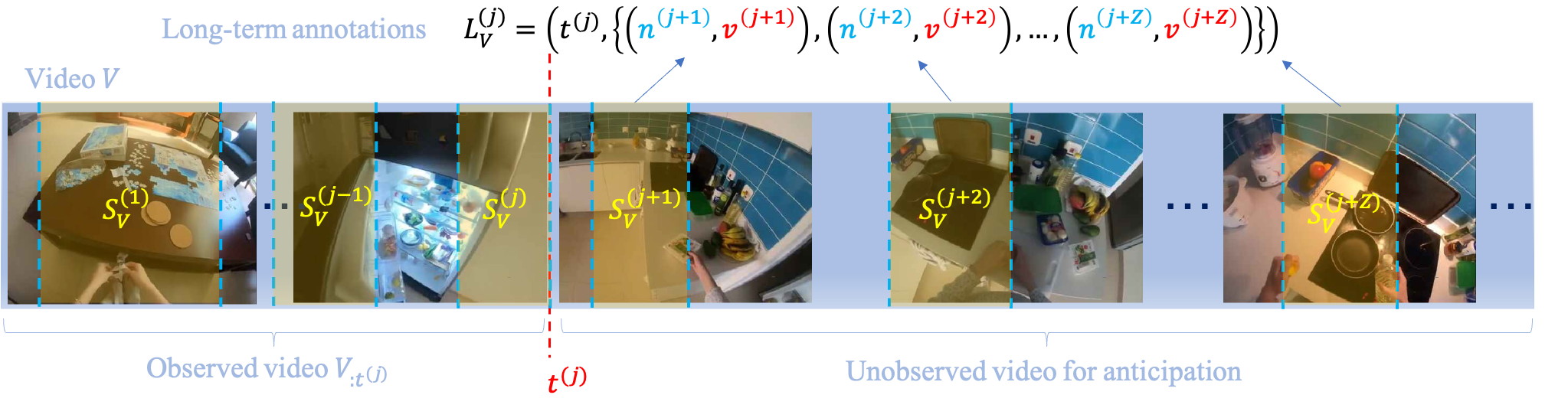}
    \label{fig:short-a}
  \end{subfigure}

\vspace{-3mm}
  \caption{\textbf{Illustration of the Long-term Action Anticipation (LTA) task.} The learned model is expected to predict a sequence of $Z$ actions, in the form of object and verb pairs, given visual observations up to time $t^{(j)}$ in the video. $t^{(j)}$ is the end time for the $j$-th labeled segment in the original video. During evaluation, the edit distance between a predicted sequence and the ground truth sequence is computed. To account for uncertainty in action anticipation, the model can predict up to $K$ sequences for each input example.}
  \label{fig:LTAtask}
\end{figure*}

\noindent\textbf{Visual-Language Pretrained Models.} We have collectively made huge progress towards building unified learning frameworks for a wide range of tasks, including language understanding~\cite{devlin2018bert, gpt2, gpt3,liu2019roberta}, visual recognition~\cite{kokkinos2017ubernet,kendall2018multi,zamir2018taskonomy,ghiasi2021multi}, and multimodal perception~\cite{jaegle2021perceiver,sun2019videobert,vilbert,girdhar2022omnivore,alayrac2022flamingo}.  

While the earlier visual-language pretrained models work with image-level~\cite{vilbert, radford2021clip} or video-level~\cite{sun2019videobert,luo2020univl} representations, more recent models are object grounded~\cite{glip,minderer2022simple,zhong2022regionclip,gu2021open,lu2022unified}. We explore the benefits of both object-level~\cite{glip} and image-level~\cite{radford2021clip} for action anticipation.

\section{Methods}
\label{sec:method}

We first introduce the long-term action anticipation (LTA) problem formulation. Then, we describe our overall model architecture. Finally, we describe our choices on object-based video representation in detail.

\begin{figure*}
  \centering
  \begin{subfigure}{.89\linewidth}
  \centering
    \includegraphics[width=0.85\linewidth]{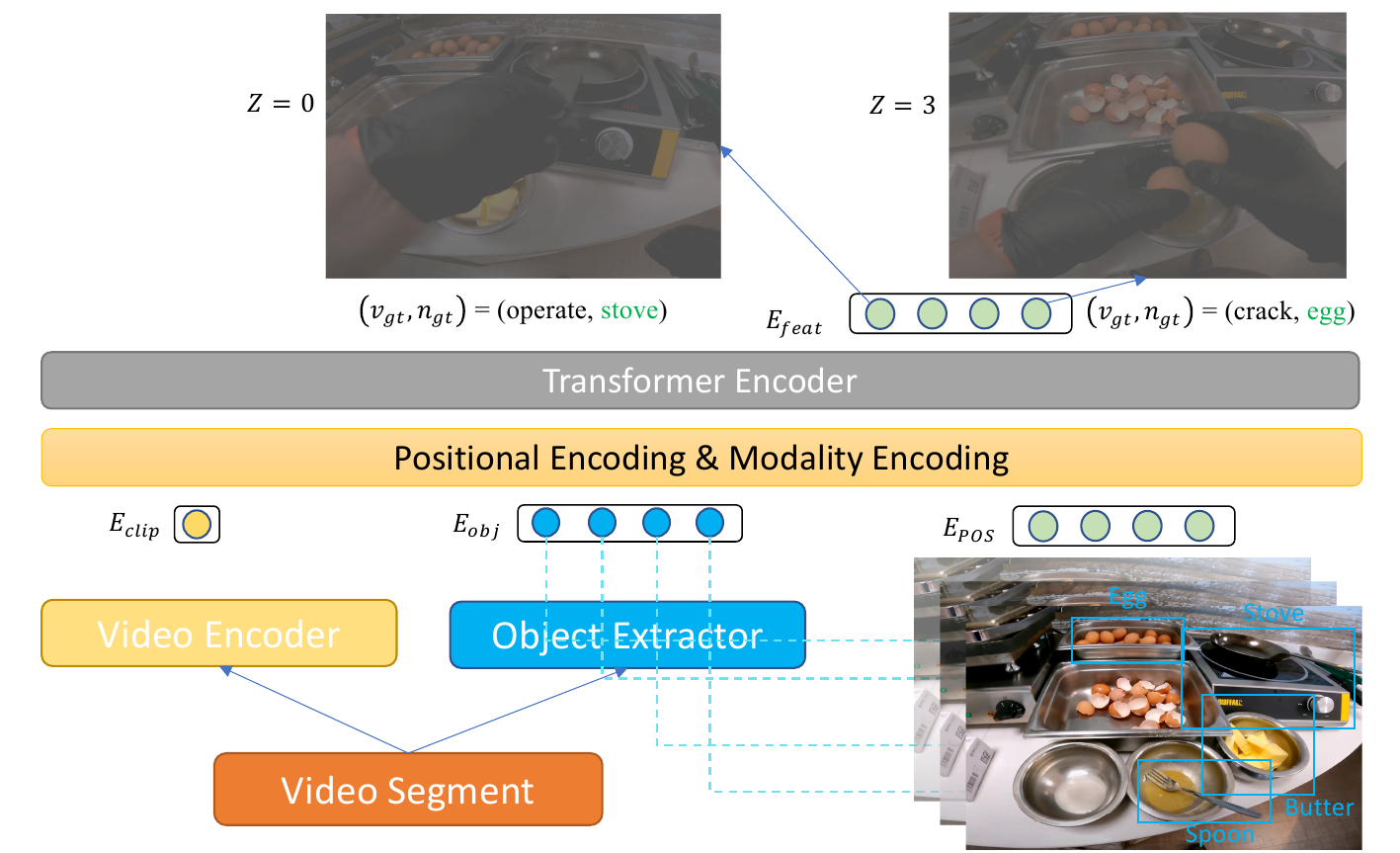}
  \end{subfigure}
  \caption{\textbf{An illustration of the overall model architecture.} We consider $N_\text{v} = 1$ input video segments, and use $N_\text{o} = 4$ objects for the video input. We consider predicting $Z = 4$ future actions. Given the features $E_\text{clip}, E_\text{obj}$ extracted by the encoders, Predictive Transformer Encoder (PTE) encodes $Z$ learnable tokens $E_\text{POS}$ to compute the features $E_\text{feat}$ for action anticipation. A decoder network is applied on each of the $Z$ encoded token $z_i$ from $E_\text{feat}$ to compute verb and object probabilities. 
  } 
  \label{fig:arch_model}
\end{figure*}

\subsection{The Action Anticipation Task}

Given the significant applications of action anticipation, numerous benchmarks have been introduced to evaluate model performance~\cite{DARKO,activityforecasting,ego-topo,Epic-Kitchen,damen2022rescaling,sun2019relational}.

We focus on the long-term action anticipation (\textbf{LTA}) problem setup introduced by Ego4D~\cite{ego4d}, whose evaluation metric takes the diversity of future prediction into account. As illustrated in Figure~\ref{fig:LTAtask}, the provided annotations first split a long video $V$ into smaller segments $\{S_V^{(i)}\}$, where $S_V^{(i)}$ is the $i$-th annotated video segment. Each segment is labeled with its starting time, end time, and action label. The action label is represented as one verb, object pair $(n^{(j)}, v^{(j)})$ for each segment $S_V^{(j)}$. The LTA task is specified by a ``stop time'' $t^{(j)}$, which denotes the end time for the last observed video segment $S_V^{(j)}$. The learned model is allowed to observe any video frames before $t^{(j)}$ in order to make future predictions $\{(n^{(j+1)}, v^{(j+1)}), ..., (n^{(j+Z)}, v^{(j+Z)})\}$, where $Z$ is the number of future steps to predict. For example, suppose a person is frying an egg with a pan in a kitchen where knives, onions, water, and pots are scattered around. In this scenario, the LTA requires the model to predict the person's upcoming actions sequentially, such as picking up the knife, cutting the onion, and drinking water. To account for the uncertainty of future behaviors, the model is allowed to make up to $K$ sets of action predictions for each future step. For Ego4D~\cite{ego4d}, the standard setup and use $Z=20$, and $K=5$. A special case is when $Z=1$, namely next action prediction. More details on the evaluation metric are in Section~\ref{sec:experiment}. 

\subsection{Overall Model Architecture}
 
Given a video $V$ and a stop time $t^{(j)}$, our model takes a sequence of video segments $\{S_V^{(j-N_\text{v}+1)}, ..., S_V^{(j)}\}$ as the input observation, and generate a sequence of actions $\{(n^{(j+1)}, v^{(j+1)}), ..., (n^{(j+Z)}, v^{(j+Z)})\}$ as outputs. $N_\text{v}$ is the number of observed video segments, and $Z$ is the number of future steps. As illustrated in Figure~\ref{fig:arch_model}, our overall model architecture consists of three modules: (1) a collection of video or object encoders that generate multimodal representations from video segments; (2) an aggregator network which fuses multimodal input representation across space and time; (3) an output decoder which generates action predictions from the aggregated features.

\vspace{.2em}\noindent\textbf{Video Representation.} We use $N_v$ clips from the observed video segments, each of which is optionally subsampled (\eg to have the same number of video frames per clip). We pass the clips to a video encoder (\eg SlowFast~\cite{slowfast}) to compute clip-level representations $E_\text{clip} \in \mathbb{R}^{N_\text{v} \times D}$, where $D$ is the encoded embedding size. We hypothesize that these clip-level representations capture scene-level appearance and motion information. We then select $N_o$ objects from the set of detected objects among the $N_v$ video segments. To compute a visual descriptor for each selected object, we apply an object encoder to generate an object embedding of the same dimension $D$. The object representation $E_\text{obj} $ has the shape of $ N_\text{o} \times D$. We discuss how to obtain the objects and their descriptors in Section~\ref{sec:object_prompt}.

\vspace{.2em}\noindent\textbf{Temporal Aggregators.}
We now introduce Predictive Transformer Encoder (PTE), a Transformer-based architecture for the action anticipation task. Given $E_\text{clip}$ and $E_\text{obj}$, PTE is used to generate $Z$ features $z_0, z_1, ..., z_{Z-1}$ for future prediction.

PTE has learnable tokens $E_\text{POS} \in \mathbb{R}^{Z \times D}$. It concatenates $E_\text{clip}$, $E_\text{obj}$ and $E_\text{POS}$ to form a sequence of length $N_\text{v} + N_\text{o} + Z$, then adds positional and modality encodings to the entire sequence. PTE uses a bidirectional Transformer encoder~\cite{transformer}, so that the learnable tokens corresponding to the future actions can be encoded in parallel. We refer to the $Z$ encoded learnable tokens as $z_0, z_1, ..., z_{Z-1}$.

\vspace{.2em}\noindent \textbf{Decoders and Training Objectives.}
For each $z_i$ encoded by PTE, we apply one set of linear classifiers on top as the decoder to compute the logits for the actions. We use two linear classifiers, one for verb, the other for noun. We apply Softmax Cross-Entropy as the loss function, and assign equal weights to all $Z$ future steps to be predicted.

\vspace{.2em}\noindent \textbf{Fusion strategies}. PTE takes video representations video encoders that capture motion, and object encoders to extract object-centric representation. These input representations can be concatenated together to form the input of PTE, which are then fused with bidirectional Transformer encoders. We refer to this approach of combining multiple video representations as early fusion. To better understand the best strategy to combine motion and object information, we also explore late fusion, where we use the two encoders to generate logits for future actions independently, and then take the average of the logits to form final predictions. 

\begin{figure*}[h!]
  \centering
  \begin{subfigure}{.8\linewidth}
  \centering
    \includegraphics[width=0.9\linewidth]{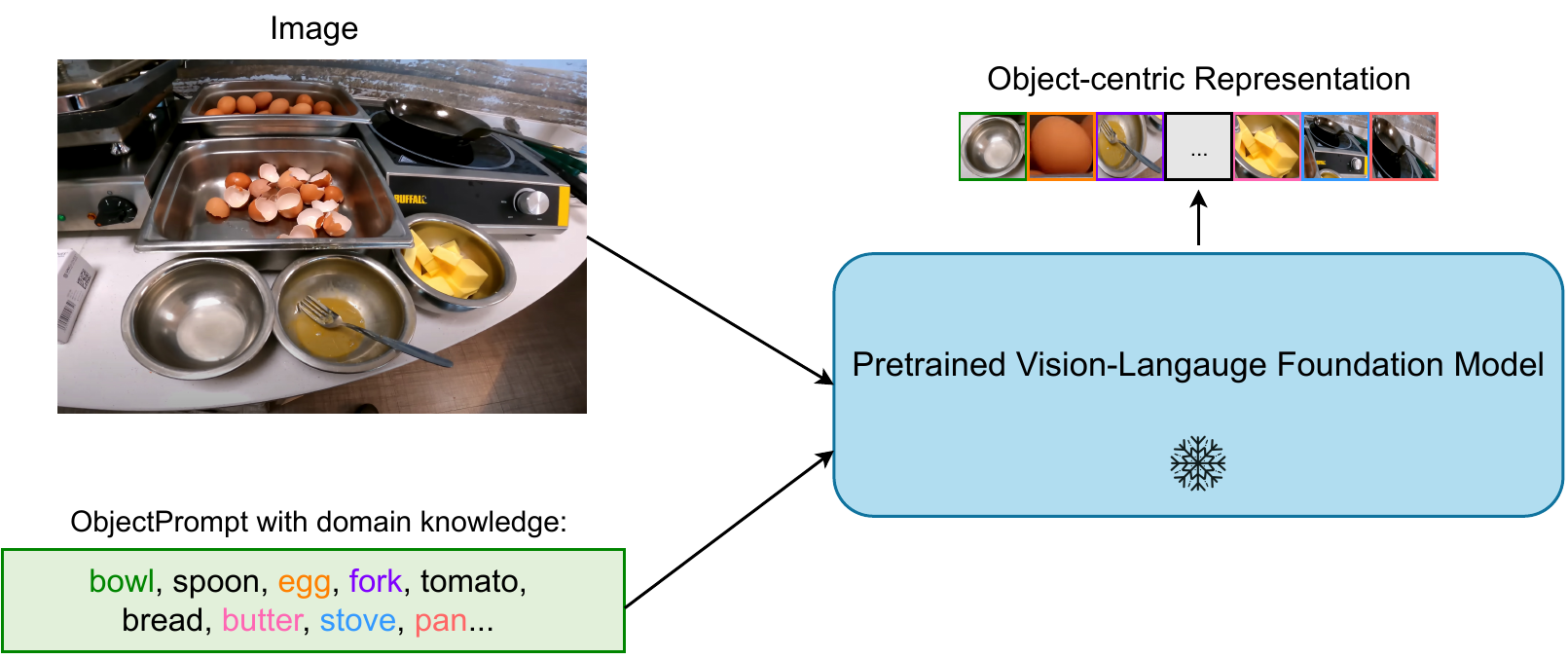}
  \end{subfigure}
   \caption{An illustration of \textbf{ObjectPrompt} to construct object-centric representation. On the left we show the input image and the list of objects of interest selected based on domain knowledge (\eg kitchen and cooking) using ObjectPrompt. The prompts are then provided to a frozen vision-language model, such as GLIP~\cite{glip}, to detect the objects of interest and extract object-centric visual representations.}
  \label{fig:object_prompt}
\end{figure*}

\subsection{Object-centric Video Representation}
\label{sec:object_prompt}

We now describe how to extract object-centric representations from videos. 

As illustrated in Figure~\ref{fig:object_prompt}, we propose to leverage a general purpose grounded vision-language model, such as GLIP~\cite{glip}, to recognize objects of interest and compute their corresponding visual descriptors. We hypothesize that models pretrained on diverse object detection and phrase grounding datasets like GLIP already encode transferrable object information, but it is crucial to \textit{guide} the model to focus on the objects of interest based on domain-knowledge (when they are available). We thus design an object prompt strategy that effectively incorporates the prior domain knowledge.

\subsubsection{Leveraging Pre-trained Grounding Models}

Our proposed framework is able to leverage any recent image-language grounding models, such as region-text contrastive models~\cite{zhong2022regionclip}, or open-vocabulary object detectors~\cite{gu2021open,minderer2022simple}. To demonstrate the effectiveness of our proposed framework, we choose one particular model Grounded Language-Image Pre-training (GLIP)~\cite{glip}, and we believe advances in grounding models would lead to more performant object-centric representations. GLIP~\cite{glip} proposes a pre-training strategy to build vision-language foundation models that can be further adopted for ``zero-shot'' object detection. During pre-training, GLIP is agnostic to the choice of object detectors to generate region proposals. In our work, we follow GLIP's setting to use Dynamic Head~\cite{DynamicHead} object detector for images, and the BERT~\cite{BERT} encoder for text inputs. Similar to applying CLIP~\cite{radford2021clip} for zero-shot image classification, the zero-shot object detection can be achieved by querying with ``object prompts'' (\eg \textit{``an image region with a cat''}) to the pre-trained language-image grounding model.

To retrieve the object descriptors computed by GLIP for the finally selected object proposals, we explore two complementary approaches. First, we inject a special identifier inside each region proposal data structure in order to gain original object features for each proposal. Second, we crop and feed the object proposal region into a pretrained image encoder to generate object features. Additionally, we also extract the object category level alignment scores and the box location to append to the object-level representation. The second approach allows us to leverage additional image-language pretrained model, such as CLIP~\cite{radford2021clip}.

\subsubsection{ObjectPrompt with Domain Knowledge}

Intuitively, domain knowledge is helpful to guide a grounding model to focus on objects of \textit{interest}. For example, for humans performing cooking activities in a kitchen (\eg Figure~\ref{fig:object_prompt}), it is important to focus on the food and cookware to model human-object interactions. Grounded pre-trained models naturally support such guidance in the form of text prompts, namely a list of objects that the model is expected to ground. We thus hypothesize that the desirable object prompts should incorporate the domain knowledge (\eg common objects appearing in the dataset), and explore different strategies to design the object prompts.

\begin{figure}[ht]
  \centering
   \includegraphics[width=\linewidth]{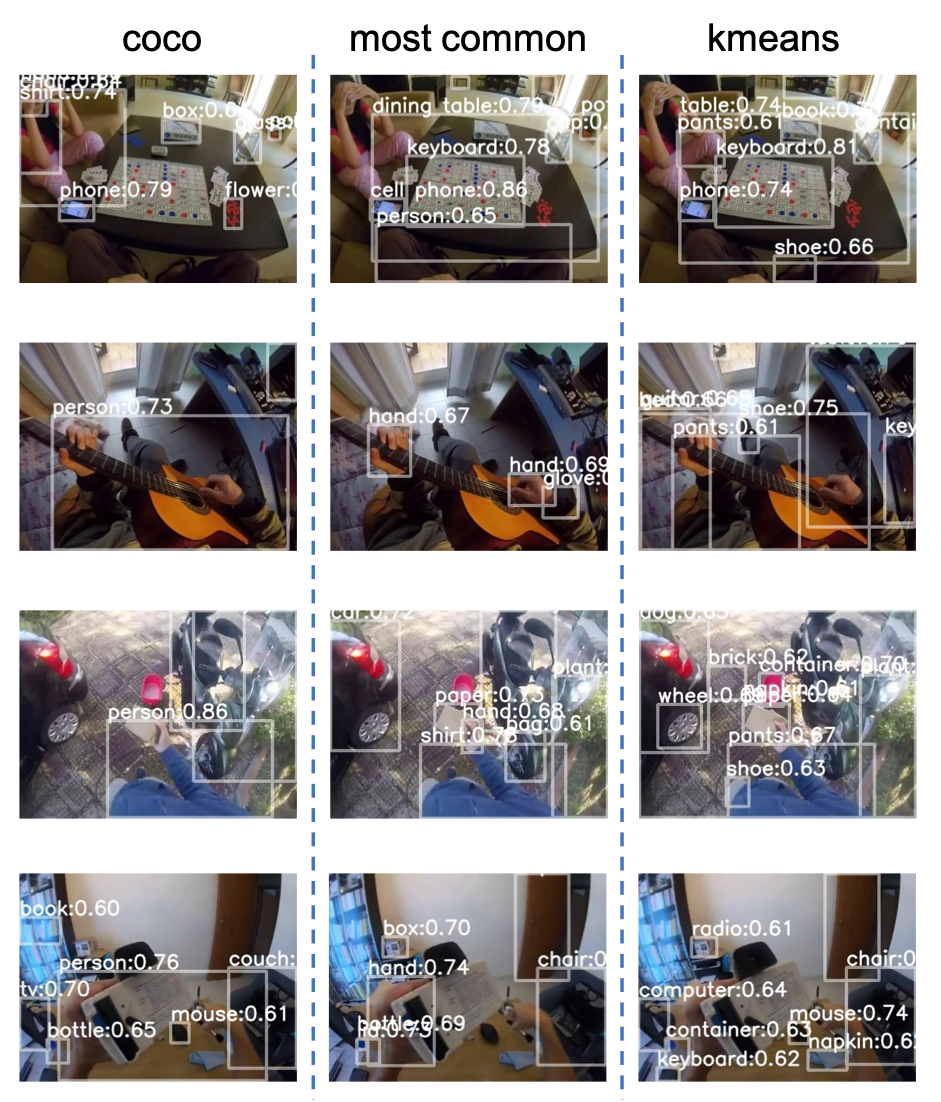}
\caption{\textbf{Example GLIP detection results with different object prompts} on randomly sampled video frames of Ego4D. We observe that the K-Means clustering strategy (last column) offers the best precision and recall for the objects of interest in all four cases.}
   \label{fig:Prompt_visualization}
\end{figure}

In order to obtain the object prompts, we first define the object vocabulary that contains the object classes to be detected. One simple and intuitive solution is to reuse the object vocabulary as provided by the task: For LTA, this refers to the list of objects (nouns) to be interacted with. To refine the vocabulary, we explore two complementary strategies: First, pick the most common object categories based on their frequency in the training data of the target task; Second, use word embedding (\eg word2vec \cite{word2vec}) and K-Means clustering to group similar categories. In addition, we also explore standard object vocabulary, such as the one used by the COCO~\cite{COCO} dataset for comparison. Figure~\ref{fig:Prompt_visualization} shows the actual detections by GLIP with different object prompt strategies. Compared to COCO, both \textit{most-common} and \textit{kmeans} are able to better detect the important objects (\eg keyboard, music instrument, and hand) thanks to the in-domain knowledge when designing object prompts.

\vspace{.2em}\noindent\textbf{Discussion.} Though well supported by research on human cognition~\cite{kahneman1992reviewing,grill2005visual,tenenbaum2011grow,10.3389/fpsyg.2016.00111,Woodward1998-WOOISE}, how to effectively construct object-centric video representations remains an open problem. For example, ObjectViViT~\cite{zhou2023can} observes that general-purpose object detectors, even capable of detecting over twenty-thousand object categories~\cite{zhou2022detecting}, do not lead to better video representation for the classification task. They instead have to resort to object detectors trained on the target video datasets, with their collected object bounding box annotations. We hypothesize that one important reason for the challenges of applying general object detectors for representing videos is the lack of attention to the objects of interest: As a result, the irrelevant objects may distract or mislead the learning framework. Instead of relying on in-domain object annotations, which are time-consuming to collect, our ObjectPrompt framework only relies on domain knowledge, which are extremely easy to collect (\eg always detect hands since it is important to model human-object interactions). In our experiments, we simply use the list of actions in a target dataset to compute the object prompts.

\section{Experiment}
\label{sec:experiment}

We now report experimental observations on three benchmarks, Ego4D, EGTEA Gaze+, and 50Salads to demonstrate the effectiveness of ObjectPrompt on action anticpation task. We perform extensive ablation study and qualitative / quantitative result analysis to support our findings. Additional implementation details can be found in the supplementary material. 

\subsection{Experiment Setup} 

\noindent \textbf{Ego4D} \cite{ego4d} contains 3,670 hours of daily life activity egocentric video spanning hundreds of scenarios. We focus on videos under Forecasting subset designed for the LTA benchmark. It contains 1723 clips with 53 scenarios and around 116 hours. The label space includes 115 verb as interaction behaviors and 478 noun as objects. We follow the standard train, validation, and test splits from \cite{ego4d} annotations for evaluation. Test split performance is obtained via the official evaluation server.

\vspace{.2em}\noindent \textbf{EGTEA Gaze+}~\cite{li2018eye} contains 28 hours egocentric videos of cooking activities from 86 unique sessions of 32 subjects. Each video is annotated with 19 verbs and 53 nouns. We follow the same train and test splits.

\vspace{.2em}\noindent \textbf{50Salads}~\cite{salad_dataset2013} contains 25 people preparing salads in 50 videos with actions falling into 17 classes. The average video length is 6 minutes and on average the number of instances per video is around 20. We follow the dataset standard and perform 5-fold cross-validation in our evaluation. 

\vspace{.2em}\noindent\textbf{Evaluation Metrics.} For Ego4D, we follow the standard evaluation protocol from~\cite{ego4d}, and report edit distance (ED). ED is computed as the Damerau-Levenshtein distance over sequences of predictions of verbs, nouns and actions. For the $K$ possible sequences a model predicts, the metric chooses the smallest edit distance between the ground truth and any of the $K$ sequences. Following \cite{ego4d}, we set $K=5$ in our experiments, and report Edit Distance at $Z = 20$ (ED@20) on the test set and Average Edit Distance (AUED) on the validation set.

For EGTEA Gaze+, we report top1 accuracy and class-mean top1 accuracy. We report next action prediction performance which is the standard practice on this dataset. For 50Salads, we follow the LTA evaluation metric in~\cite{lta_seg} and report the mean over classes accuracy (MoC@($O$,$F$)), where $O$ is the percentage of observed video frames, and $F$ is the percentage of future video frames to be labeled.

\begin{table*}[h]
\centering
\subfloat[\textbf{Object Prompt Strategy}\label{ego4d_vocabulary}]{
\vspace{-1em}
\hspace{-1em}
    \begin{tabular}{@{}ccc@{}}
    \toprule
    Vocabulary &  Verb $\downarrow$  & Noun $\downarrow$ \\ 
    \midrule
    random bbox & 0.745 & 0.945 \\
    coco & 0.737 (-1.0\%) & 0.789 (-16.5\%) \\
    most common & 0.734 (-1.5\%) & 0.776 (-17.8\%) \\
    kmeans & \textbf{0.728 (-2.3\%)} & \textbf{0.771 (-18.4\%)}\\
    \bottomrule
    \label{tab:ablation_vocab}
    \end{tabular}
}
\subfloat[\textbf{Number of Objects}\label{ego4d_objects}]{
\vspace{-1em}
\hspace{1em}
    \begin{tabular}{@{}ccc@{}}
    \toprule
    \#obj & Verb $\downarrow$  & Noun $\downarrow$ \\ 
    \midrule
    1 & 0.735 & 0.834 \\
    3 & 0.727 & 0.800 \\
    5 & \textbf{0.728} & \textbf{0.771} \\
    10 & 0.731 & 0.781 \\
    \bottomrule
    \label{tab:ablation_num_objs}
    \end{tabular}
} \hfill
\subfloat[\textbf{Location and Category}\label{ego4d_loc_cat}]{
\vspace{-1em}
\hspace{-1em}
    \begin{tabular}{@{}cccc@{}}
    \toprule
    Loc. & Cate. & Verb$ \downarrow$  & Noun$ \downarrow$ \\ 
    \midrule
    \xmark & \xmark & 0.730 & 0.789 \\
    \cmark & \xmark & 0.730 & 0.786 \\
    \xmark & \cmark & 0.731 & 0.774 \\
    \cmark & \cmark & \textbf{0.728} & \textbf{0.771} \\
    \bottomrule
    \label{tab:ablation_loc_cat}
    \end{tabular}
}
\subfloat[\textbf{Threshold}\label{ego4d_threshold}]{
\vspace{-1em}
\hspace{1em}
    \begin{tabular}{@{}ccc@{}}
    \toprule
    Threshold & Verb $\downarrow$  & Noun $\downarrow$ \\ 
    \midrule
    0.00 & 0.731 & 0.781 \\
    0.30 & \textbf{0.728} & \textbf{0.771} \\
    0.45 & 0.728 & 0.773 \\
    0.55 & 0.731 & 0.787\\
    \bottomrule
    \label{tab:ablation_thre}
    \end{tabular}
}
    
\caption{\textbf{Ablation study of object-centric representation on the Ego4D validation set}. We conduct detailed ablation on (1) object vocabulary, (2) number of object per frame, (3) object location and category, (4) detection threshold. We only provide object-centric representation $E_\text{obj}$ as the video representation.}
\label{tab:ablation_object_only}

\end{table*}

\subsection{Ablation Study}
\label{subsec:ablations}
We focus our ablation study on the Ego4D dataset. To better understand the effectiveness of our object-centric representation, we only provide the object features $E_\text{obj}$ as the input to the PTE model.

\vspace{.2em}\noindent\textbf{Object prompts.} We first study how to properly incorporate domain knowledge into object prompt design. We explore three strategies: ``most-common'', ``kmeans'', ``coco''.
We fix the number of prompts to contain 80 object prompts, and empirically observe that the performance is robust even with fewer (\eg 20 and 40) prompts. As a recap, the ``most-common'' strategy pickes the top 80 most frequent nouns from the Ego4D vocabulary. The ``kmeans'' strategy contains the top 80 most frequent prompts after the entire list of objects is clustered to reduce redundancy. The ``coco'' baseline uses the 80 object categories from the COCO~\cite{COCO} dataset. We also include ``random'' as an additional baseline. It randomly picks regions in images as bounding boxes. From Table~\ref{tab:ablation_vocab}, we can see that both ``most-common'' and ``kmeans'' outperform the baselines ``coco'' and ``random'' significantly, demonstrating the effectiveness and importance of our designed object prompts.

\vspace{.2em}\noindent\textbf{Object locations and categories.} In Table~\ref{tab:ablation_loc_cat}, we observe that both location and category features provide useful information when included in the object-centric representation, especially for the noun ED. We also observe that object location information appears to be less important than category information. We conjecture that the category information may provide the model with high-level scene information (\eg \textit{person}, \textit{laptop}, and \textit{book} is very likely to indicate a library), which helps the model infer long-term future actions.

\vspace{.2em}\noindent\textbf{Object quantity and quality.} Each grounded object by the GLIP model comes with a confidence score. Intuitively, higher scores correspond to higher quality objects, at the cost of lower recall. In Table~\ref{tab:ablation_thre}, we use a threshold on confidence scores to filter objects.
We can see that the model performs the best when we set threshold to 0.3. In Table~\ref{tab:ablation_num_objs}, we vary the maximum number of objects to be selected per frame, sorted via confidence scores. We can see the model performs the best when we use 5 objects per frame. We conjecture that a trade-off exists between the quantity and quality of objects. 

\vspace{.2em}\noindent\textbf{Temporal modeling.} We compare our temporal modeling methods with Ego4D LTA Baseline~\cite{ego4d} in the first two rows of Table \ref{tab:ablation_ego4d}. We observe that with the same video representation $E_\text{clip}$ encoded by a pre-trained SlowFast network, PTE achieves significant improvement over the baseline temporal aggregator with a vanilla Transformer with no learnable tokens for future prediction.

\vspace{.2em}\noindent\textbf{Effectiveness of the object-centric representation.} The next two rows of Table~\ref{tab:ablation_ego4d} study the impact of object-centric representation. We observe that the object modality helps reduce the edit distance substantially, especially for noun prediction. We also observe that early fusion works better than late fusion.

\begin{table}[t]
\centering
\setlength{\tabcolsep}{1pt}
\scalebox{.9}{
\begin{tabular}{@{}ccccc@{}}
\toprule
Aggregator & Modality & Fusion & Verb $\downarrow$ & Noun $\downarrow$ \\ 
\midrule
Transformer & video & - & 0.751 & 0.766 \\
PTE & video & - & 0.713 (-5.1\%) & 0.753 (-1.7\%) \\
PTE & video+object & early & \textbf{0.707(-5.9\%)} & \textbf{0.743 (-3.0\%)} \\
PTE & video+object & late & 0.709 (-5.6\%) & 0.748 (-2.3\%) \\
\bottomrule
\end{tabular}
}
\caption{\textbf{Temporal modeling and modality fusion on Ego4D LTA validation set.} We observe that PTE outperforms the baseline temporal aggregator, and object-centric representation can effectively improve the performance, especially with early fusion.}
\label{tab:ablation_ego4d}
\end{table}

\vspace{.2em}\noindent\textbf{Enhancing object-centric representation.}
While we use GLIP \cite{glip} to detect and represent objects, their are many other pre-trained models we can use to obtain object representation. In Table~\ref{tab:ablation_pretrained_models}, we explore additional pretrained models for object representation. We still use the same detected objects, and use CLIP~\cite{radford2021clip} (ViT-L/14@336px) to represent detected objects. We observe that incorporating CLIP representation further reduces the edit distance. An interesting future work is to explore other pre-trained visual encoders.

\begin{table}[t]
\centering
\setlength{\tabcolsep}{3pt}
\begin{tabular}{@{}cccc@{}}
\toprule
Model & Modality & Verb$\downarrow$ & Noun$\downarrow$ \\ 
\midrule
- & video & 0.713 & 0.753 \\
GLIP & video+object & 0.707 (-5.9\%) & 0.743 (-3.0\%) \\
CLIP & video+object & \textbf{0.700 (-6.8\%)} & \textbf{0.717 (-4.8\%)} \\
\bottomrule
\end{tabular}
\caption{\textbf{Impact of object-centric representation on Ego4D LTA validation set.} Incorporating CLIP brings additional performance gain over GLIP object embeddings.}
\label{tab:ablation_pretrained_models}
\end{table}

\begin{figure}[h]
  \centering
   \includegraphics[width=\linewidth]{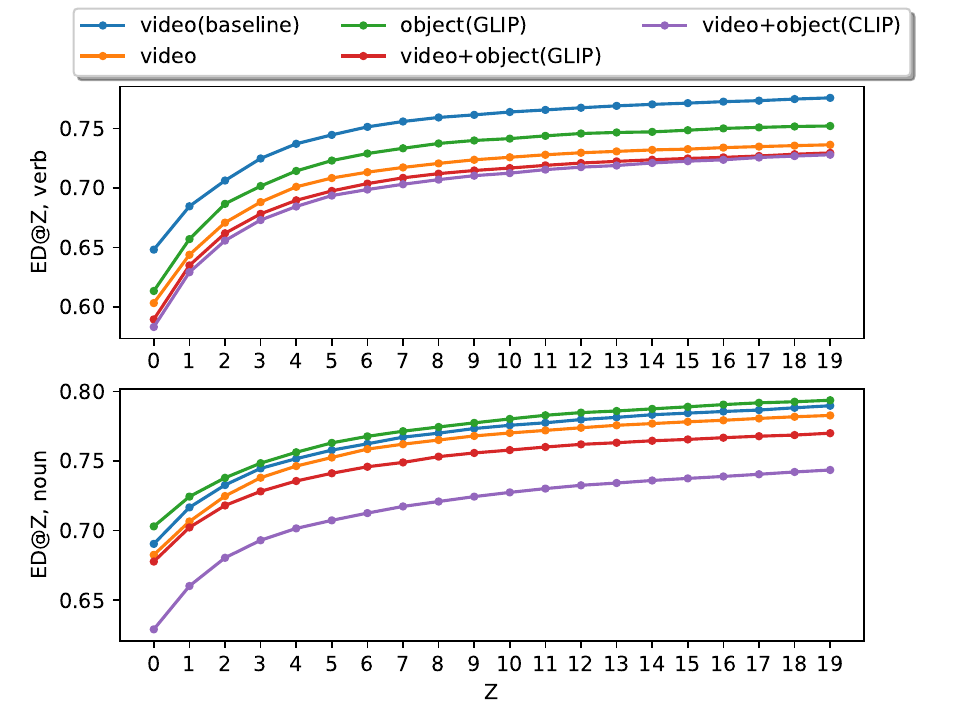}
   \caption{\textbf{Per-step ED} on Ego4D validation (lower is better). \textbf{Top}: verb. \textbf{Bottom}: noun. ED@Z is the edit distance at $Z^{th}$ step.}
   \label{fig:ed-per-step}
\end{figure}

\vspace{.2em}\noindent\textbf{Per-step analysis.}
We further analyze the per-step edit distance for verbs and nouns on the Ego4D validation set in Figure \ref{fig:ed-per-step}. ``video (baseline)'' uses the vanilla Transformer Encoder as temporal aggregator. The other methods use PTE as the temporal aggregator. We can see that our object-centric representation leads to consistent and significant improvements across all 20 steps, especially for the nouns.

\vspace{-0.5em}
\begin{table}
\setlength{\tabcolsep}{3pt}
\centering
    \begin{tabular}{@{}cccc@{}}
    \toprule
    Model & Verb $\downarrow$ & Noun $\downarrow$ & Action $\downarrow$ \\ 
    \midrule
    ICVAE~\cite{icvae} & 0.7410 & 0.7396 & 0.9304 \\
    SlowFast~\cite{ego4d} & 0.7389 & 0.7800 & 0.9432 \\
    VCLIP~\cite{das2022video+} & 0.7389 & 0.7688 & 0.9412 \\
    \midrule
    ObjectPrompt (ours) & \textbf{0.7265} & \textbf{0.7396} & \textbf{0.9290} \\
    \bottomrule
    \end{tabular}
    \caption{Comparison with the state-of-the-art on the Ego4D LTA benchmark. We report test split results from the evaluation server.}
    \label{sota:ego4d}
\end{table}

\begin{table}
\centering
\scalebox{.7}{
    \begin{tabular}{c|cccc|cccc}
    \toprule
    Observation $\rightarrow$ & \multicolumn{4}{c|}{20\%} & \multicolumn{4}{c}{30\%}\\
Future $\rightarrow$ & 10\% & 20\% & 30\% & 50\% & 10\% & 20\% & 30\% & 50\% \\
    \midrule
    RNN~\cite{lta_seg} & 30.1 & 25.4 & 18.7 & 13.5 & 30.8 & 17.2 & 14.8 & 9.8 \\
    CNN~\cite{lta_seg} & 21.2 & 19.0 & 15.9 & 9.8 & 29.1 & 20.1 & 17.5 & 10.9 \\
    Sener \etal~\cite{actionbanks} & 34.7 & 26.3 & 23.7 & 15.7 & 34.5 & 26.1 & 22.7 & 17.1 \\
    Qi \etal~\cite{qi2021self} & 37.9 & 28.8 & 21.3 & 11.1 & \textbf{37.5} & 24.1 & 17.1 & 9.1 \\
    FUTR~\cite{gong2022future} & \textbf{39.6} & 27.5 & 23.3 & 17.8 & 35.2 & \textbf{24.9} & 24.2 & 15.3 \\
    \midrule
    ObjectPrompt (ours) & 37.4 & \textbf{28.9} & \textbf{24.2} & \textbf{18.1} & 28.0 & 24.0 & \textbf{24.3} & \textbf{19.3}\\
    \bottomrule
    \end{tabular}
}
    \caption{Comparison with state-of-the-art on 50Salads LTA task.}
    \vspace{-.5em}
    \label{salads}
\end{table}

\begin{table}
\centering
\scalebox{.9}{
    \begin{tabular}{@{}ccc@{}}
        \toprule
        Model & Top-1 Acc. $\uparrow$ & Class-mean Acc. $\uparrow$\\ 
        \midrule
        I3D-Res50~\cite{kay2017kinetics} & 34.8 & 23.2 \\
        FHOI~\cite{fhoi}  & 36.6 & 25.3 \\
        AVT~\cite{girdhar2021anticipative} & 43.0 & 35.2 \\
        \midrule
        ObjectPrompt (ours) & \textbf{44.6} & \textbf{36.4} \\
        \bottomrule
    \end{tabular}
    }
    \caption{Comparison with state-of-the-art on EGTEA Gaze+.}
    \vspace{-.5em}
    \label{sota:gaze}
\end{table}

\subsection{Comparison to the State-of-the-art}
\label{subsec:comparison}

Finally, we report the action anticipation performance on Ego4D, 50Salads, and EGTEA Gaze+ benchmarks, and compare with previously published results.

\vspace{.2em}\noindent\textbf{Ego4D.}   Table~\ref{sota:ego4d} compares our best model (Slowfast + PTE (video + CLIP object, early)) with the recent state-of-the-art methods under comparable settings. We report verb, noun and action ED on the test set. Our model is able to outperform competitive baseline methods.

\vspace{.2em}\noindent\textbf{50Salads.}
We conduct experiments on long-term action anticipation benchmarks and report class-mean top1 action accuracy in Table \ref{salads}. Following common practice \cite{lta_seg}, we predict 10\%, 20\%, 30\%, and 50\%, respectively, of the video after observing the previous 20\% or 30\%. We use PTE as a temporal modeling method in video-only models and CLIP object representation in video+object models. Our method achieves overall on par performance with FUTR\cite{gong2022future} and achieves the best performance for longer-term anticipation.

\vspace{.2em}\noindent\textbf{EGTEA Gaze+.} 
Table \ref{sota:gaze} shows results on the next action prediction benchmarks for EGTEA+ Split 1 following the standard practice. We use CLIP object representation in video+object models, PTE as temporal modeling, and finetune the video backbone. We report top1 accuracy and class-mean top1 accuracy. 

\subsection{Qualitative Analysis}
\label{subsec:qualitative_analysis}
\begin{figure}[htp]
  \centering
  \vspace{-1em}
   \includegraphics[width=.95\linewidth]{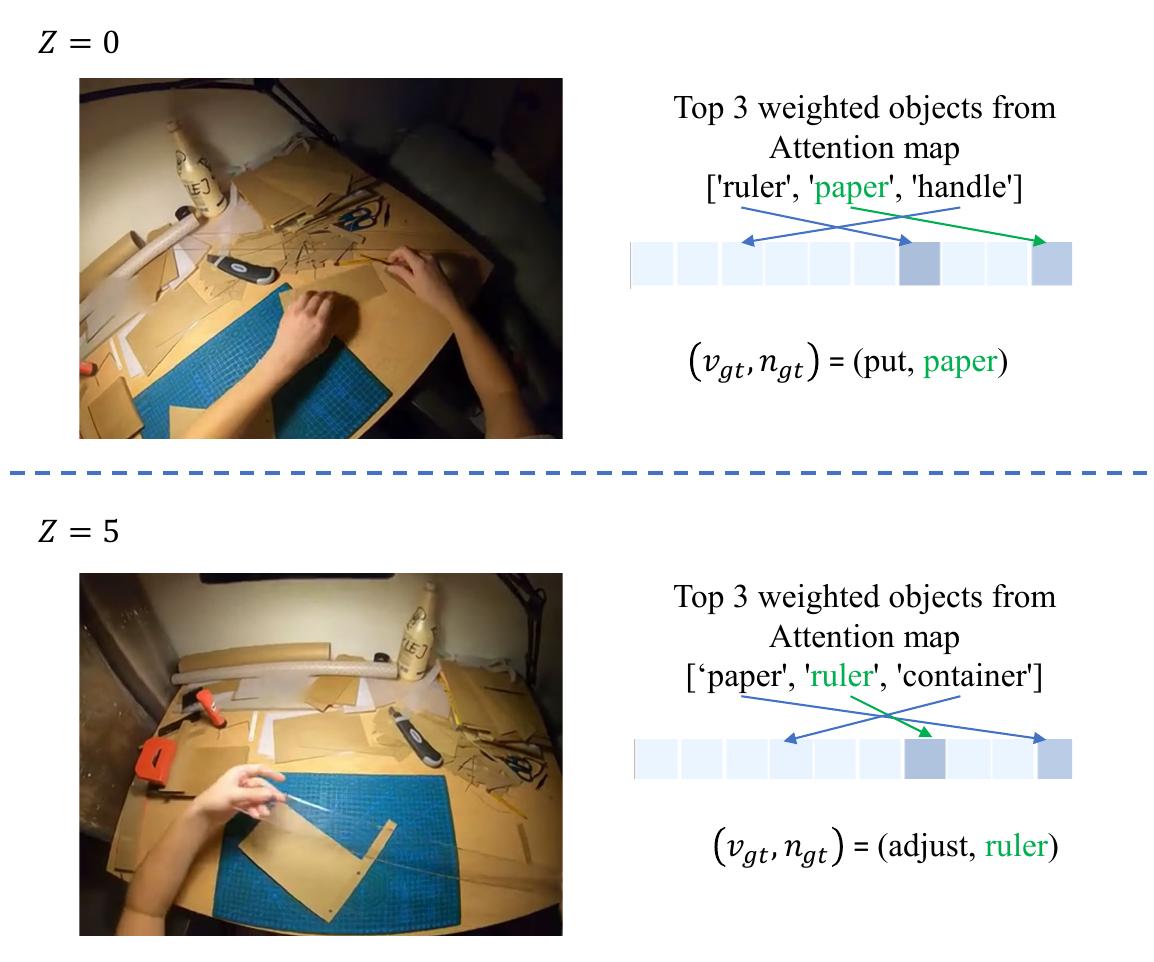}
   \caption{\textbf{Object attention visualizations.} We show the top 3 predicted objects generated with attention rollout~\cite{attentionrollout}.}
   \vspace{-1em}
   \label{fig:visual_objects}
\end{figure}

In Fig.~\ref{fig:visual_objects}, we show several qualitative examples of object attention weights produced by PTE. We use attention rollout \cite{attentionrollout} to compute attention weights from $Z$ output action (noun, verb) pairs to previous visual observations and choose the top 10 objects which have the relatively highest weight. Comparing with the ground truth label shows the model
learns to associate the corresponding objects when predicting actions at different time steps. More results can be found in the supplementary materials.

\section{Conclusion}

We propose ObjectPrompt, a framework to effectively construct object-centric video representation for long-term action anticipation. Our framework has two modules, an object prompt strategy to effectively leverage domain knowledge and construct object-centric representation with pre-trained gounding models, and predictive transformer encoder, which dynamically associates the object evidence for long time-horizon action prediction. Experiments results confirm that both modules improve the action anticipation performance. We report competitive results on Ego4D LTA, 50Salads, and EGTEA Gaze+ benchmarks.

\noindent\textbf{Acknowledgements.} This work is in part supported by Honda Research Institute and Meta AI.

{\small
\bibliographystyle{ieee_fullname}
\bibliography{main}
}

\clearpage
\newpage
\appendix
\setcounter{table}{0}
\renewcommand{\thetable}{A\arabic{table}}
\setcounter{figure}{0}
\renewcommand{\thefigure}{A\arabic{figure}}

\section{Implementation details}
We now provide additional implementation details.
\subsection{Video Data Preprocessing}
We downsample the original videos to 320p, then normalize with mean = 0.45, std = 0.225 for all pixels and channels. During training, we randomly scale the short side between 256 and 320px keeping the original aspect ratio, and then randomly crop a $224\times 224$px region from the resized image. During test, we scale the video to 256px, then take the center $224\times 224$px crop as inputs. We use Slowfast(8x8, R101) backbone provided by \cite{ego4d}. The video representation dimension is 2048 for each video segment.

\vspace{.2em}\noindent\textbf{Ego4D} provides video segment annotations in the form of start and end time stamps. The longest segments are around 8 seconds. To encode each video segment, we randomly pick a window of $128$ frames and then subsample the frames to $64$ with a stride of $2$. We set the number of observed video segments $N_v = 3$. Using more observed video segments is left as a future work. We follow the standard baseline implementation in Ego4D~\cite{ego4d}, and use SlowFast video backbone pretrained on the Kinetics-400 classification and Ego4D recognition tasks. The video backbone is frozen when transferred to the LTA task.

\vspace{.2em}\noindent\textbf{50Salads} does not provide the start and end timestamps for the observed video segments. We thus set the segment length to be 3 seconds and take $N_v = 3$ consecutive segments as inputs, resulting in 9-second video observations for both the next action prediction and the long-term action anticipation tasks. For the NAP task, we follow the standard evaluation protocol, and sample the observed video segments up till 1 second before the start of the future action. For the LTA task, the segments are sampled right before the start of the future. We sample 64 frames from each 3-second video segment with stride of 2. We use the SlowFast video backbone pretrained on the Kinetics-400 classification task. The video backbone is fine-tuned during transfer learning.

\vspace{.2em}\noindent For \textbf{EGTEA Gaze+},
we also sample $N_v = 3$ consecutive segments of 3-second each as inputs, up till 0.5 second before the start of the future action following the standard practice. We use the SlowFast video backbone pretrained on the Kinetics-400 classification task. The video backbone is fine-tuned during transfer learning.

\subsection{Object Detection}
We uniformly sample $N_\text{img}$ frames from each video segment, then take the top $N_\text{obj}$ objects from each frame based on their confidence scores. Therefore, the total number of observed object tokens $N_{o} = N_{v} \cdot N_\text{img} \cdot N_\text{obj}$. We treat the whole frame as an object so that all sampled frames have at least one object token.
For Ego4D, we set $N_\text{img} = 4, N_\text{obj} = 11$. For 50Salads and EGTEA Gaze+, we set $N_\text{img} = 2, N_\text{obj} = 11$, since their video segments are shorter.

\subsection{Object Representation}

We present the details to retrieve the object representation from visual-language pretrained models as follows.

\subsubsection{GLIP-based Object Representation}
We use the anchor-free object detection model provided by GLIP.
To retrieve the corresponding object-centric representations for the detected objects, we inject a special identifier inside each region proposal data structure after the initial region proposal generation stage (for one-stage object detectors, these correspond to different locations on a feature map, or a feature pyramid~\cite{fpn}), which contains index information corresponding to the original features. We then retrieve the original feature descriptors of the detected objects by back-tracing the identifier index information. We use the same feature embedding to compute object-text alignment scores as the representation of a detected object.

\subsubsection{CLIP-based Object Representation}

We adopt a two-stage pipeline by first detecting and selecting the objects with GLIP, and then extracting visual descriptors for each detected region. We grasp each detection region's axis indices in xyxy format, which corresponds to the left-hand side of the region, the top of the region, the right of the region, and the bottom of the region. Then we compute the center axis indices of each detection region and select the longer side between height and width as the output region side length. After that, we crop a square-like output region from the origin video frame based on center axis indices and corresponding side lengths and then feed the set of output object proposal regions into a pretrained image encoder to generate object features. Finally, before putting generated object features into PTE, we perform image normalization following~\cite{radford2021clip}.

\subsection{Predictive Transformer Encoder}
For Ego4D, PTE uses 3 Transformer blocks for object-only and video-only models, 5 blocks for video+object models. For 50Salads and EGTEA Gaze+, PTE uses 3 Tranformer blocks for all models. The number of blocks are chosen by the performance on the validation set. All transformer blocks use 8 heads. Dropout rate is 0.1. \\
In each training example, we use $N_{v}$ input video segments, $N_{o} = N_{v} \cdot N_\text{img} \cdot N_\text{obj}$ objects and $Z$ learnable prediction tokens in total. To generate objects, we first sample $N_{img}$ frames from each segments, then sample $N_{obj}$ objects 
 from each frame. Since the input sequences include segment-level and frame-level tokens, we use both segment-level and frame-level positional encoding. We add the segment-level and frame-level positional encodings to the original input sequences to form the final positional encoded inputs.
 
\vspace{.2em}\noindent \textbf{Segment-level positional encoding} is based on sinusoidal positional encodings~\cite{transformer}. Video and object tokens that are in the same segments have the same segment-level positional encoding. We also encode the learnable prediction tokens with segment-level positional encoding.

\vspace{.2em}\noindent\textbf{Frame-level positional encoding} is based on learnable positional encoding initialized with zeros. Videos and learnable prediction tokens are segment-level tokens, so we do not apply frame-level positional encoding on them. Objects that are sampled from the same segments have the same frame-level encoding.

\vspace{.2em}\noindent\textbf{Modality Encoding.} We also use learnable modality (token) type encodings for videos, objects, and also the learnable prediction tokens. They are initialized with normal distribution. The modality type encodings are added to the corresponding token representations.

\subsection{Training details}
We train our model on 8 NVIDIA GeForce RTX 3090 GPUs or 4 NVIDIA RTX A6000 GPUs with batchsize 32. We use SGD + Nesterov Momentum Optimizer. We always use the first 3 epochs for warmup and remained epochs for cosine annealed decay. We train our model for 40 epochs for Ego4D, 60 epochs for 50Salads, 20 epochs for EGTEA Gaze+. 
For Ego4D, we set learning rate to $5\times 10^{-4}$, weight decay to $1\times 10^{-4}$. For 50Salads and EGTEA Gaze+, we set learning rate to $0.02$. We apply element-wise dropout for all tokens after positional encoding and modality encoding. Dropout rate is 0.5 for the Ego4D dataset, 0.2 for 50 Salads and EGTEA Gaze+. Additionally, we apply DropToken~\cite{akbari2021vatt} with dropout rate 0.5 to apply additional regularization on the object representations.
All hyperparameters are selected based on the validation performance. 

\section{Additional Analysis}

\subsection{Object Attention Visualizations}

\begin{figure*}
  \centering
  \begin{subfigure}{\linewidth}
  \centering
    \includegraphics[width=\linewidth,page=1]{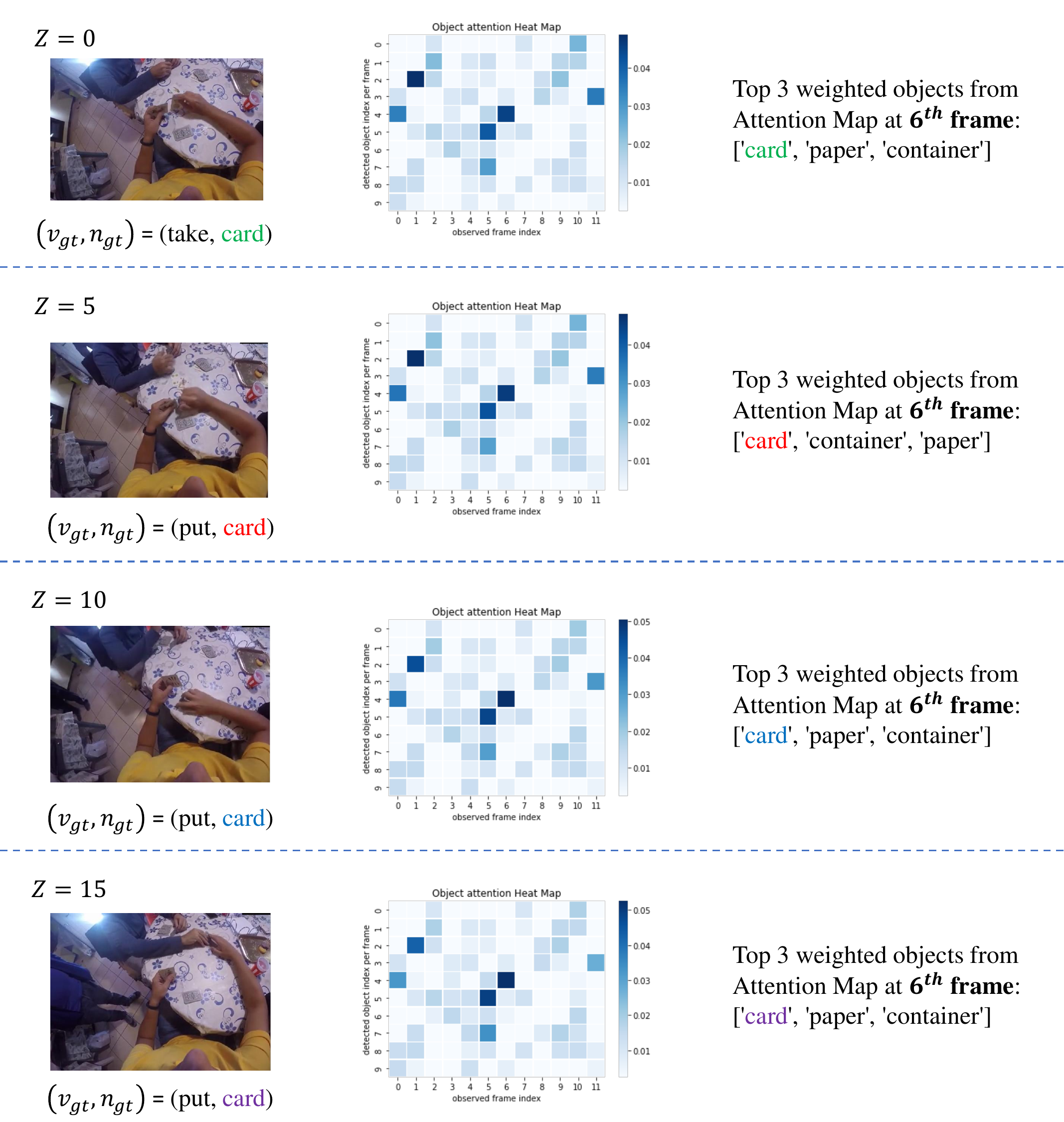}
  \end{subfigure}
  \caption{\textbf{Visualization of object attention heatmap and retrieved objects }, when $N_\text{seg}=3, N_\text{img}=4, N_\text{obj}=10, Z = 20$. \textbf{Left}: Representative video frame 
  correlate to $Z^{th}$ future step for anticipation, where $(v_{gt}, n_{gt})$ are ground truth action labels at $Z^{th}$ step in $(\text{verb}, \text{noun})$ pairs. \textbf{Middle}: Normalized object attention heatmap to previous observed visual input at $Z^{th}$ step. \textbf{Right}: Top 3 weighted objects from heatmap according to the center observed ($6^{th}$) frame.}
    \vspace{5em}
  \label{fig:heatmap1}
\end{figure*}

\begin{figure*}
  \centering
  \begin{subfigure}{\linewidth}
  \centering
    \includegraphics[width=\linewidth,page=4]{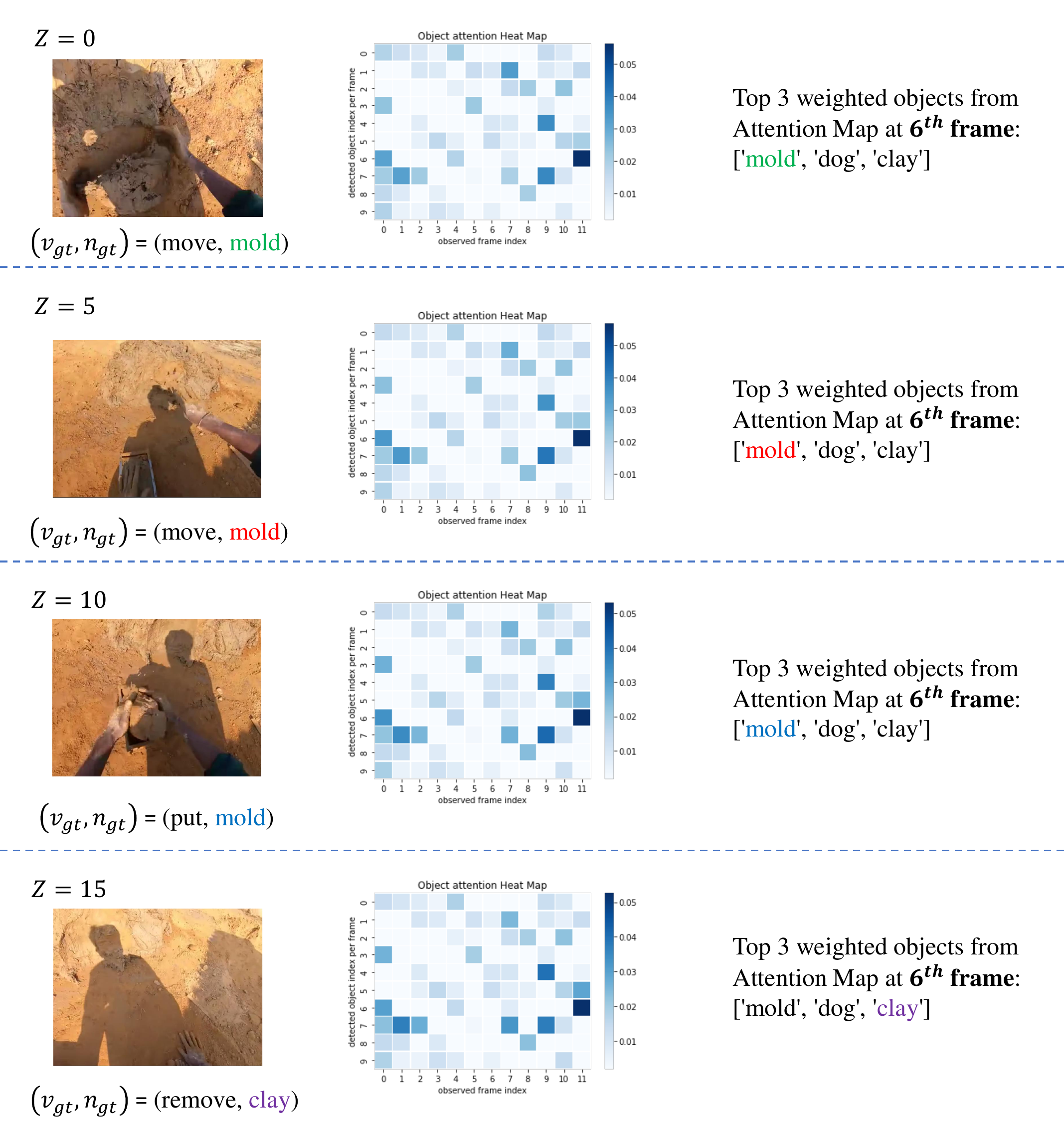}
  \end{subfigure}
  \caption{\textbf{Visualization of object attention heatmap and retrieved objects }, when $N_\text{seg}=3, N_\text{img}=4, N_\text{obj}=10, Z = 20$. \textbf{Left}: Representative video frame 
  correlate to $Z^{th}$ future step for anticipation, where $(v_{gt}, n_{gt})$ are ground truth action labels at $Z^{th}$ step in $(\text{verb}, \text{noun})$ pairs. \textbf{Middle}: Normalized object attention heatmap to previous observed visual input at $Z^{th}$ step. \textbf{Right}: Top 3 weighted objects from heatmap according to the center observed ($6^{th}$) frame.}
    \vspace{5em}

  \label{fig:heatmap4}
\end{figure*}

\begin{figure*}
  \centering
  \begin{subfigure}{\linewidth}
  \centering
    \includegraphics[width=\linewidth,page=5]{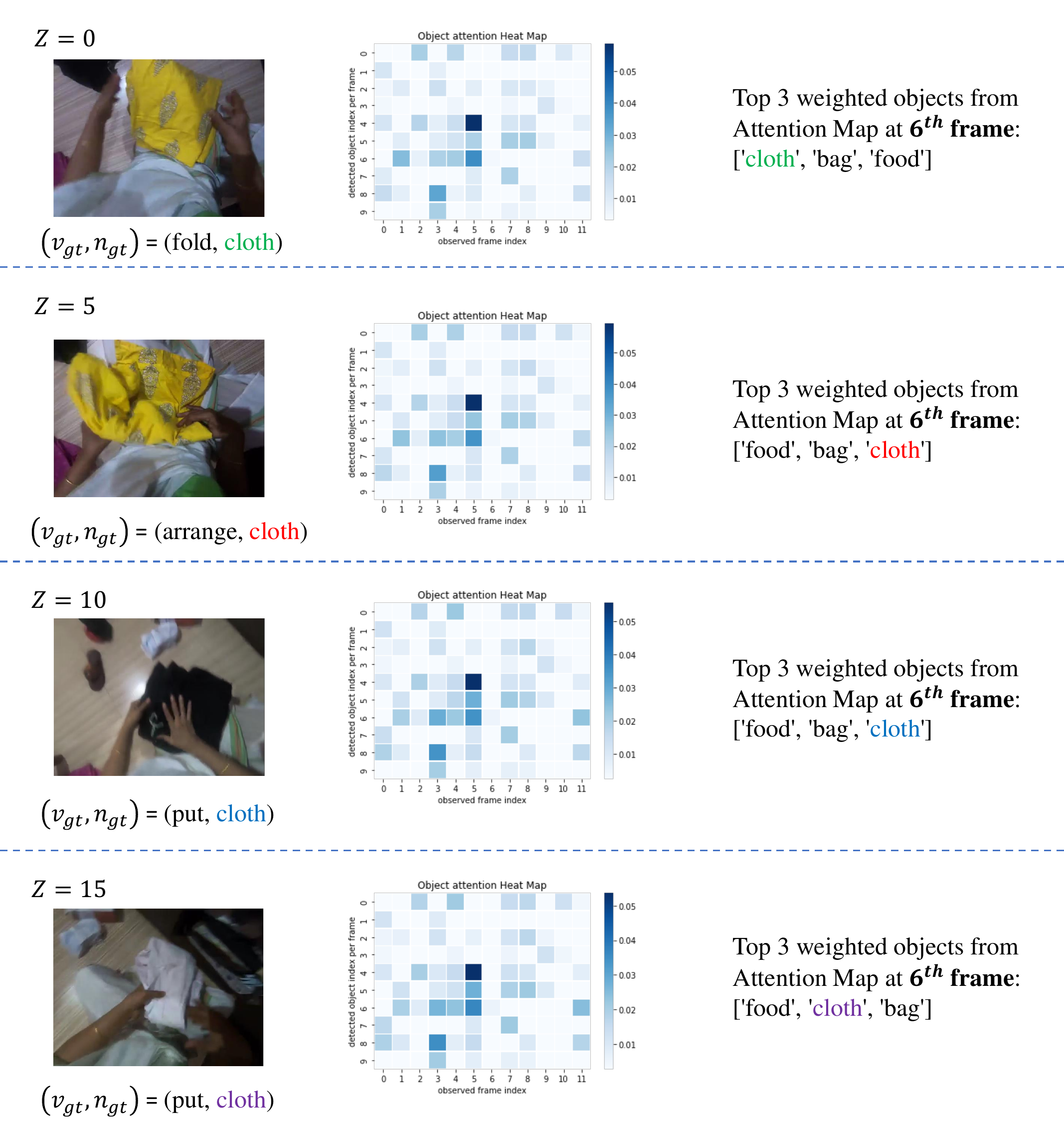}
  \end{subfigure}
  \caption{\textbf{Visualization of object attention heatmap and retrieved objects }, when $N_\text{seg}=3, N_\text{img}=4, N_\text{obj}=10, Z = 20$. \textbf{Left}: Representative video frame 
  correlate to $Z^{th}$ future step for anticipation, where $(v_{gt}, n_{gt})$ are ground truth action labels at $Z^{th}$ step in $(\text{verb}, \text{noun})$ pairs. \textbf{Middle}: Normalized object attention heatmap to previous observed visual input at $Z^{th}$ step. \textbf{Right}: Top 3 weighted objects from heatmap according to the center observed ($6^{th}$) frame.}
    \vspace{5em}
  \label{fig:heatmap5}
\end{figure*}

\begin{figure*}
  \centering
  \begin{subfigure}{\linewidth}
  \centering
    \includegraphics[width=\linewidth,page=6]{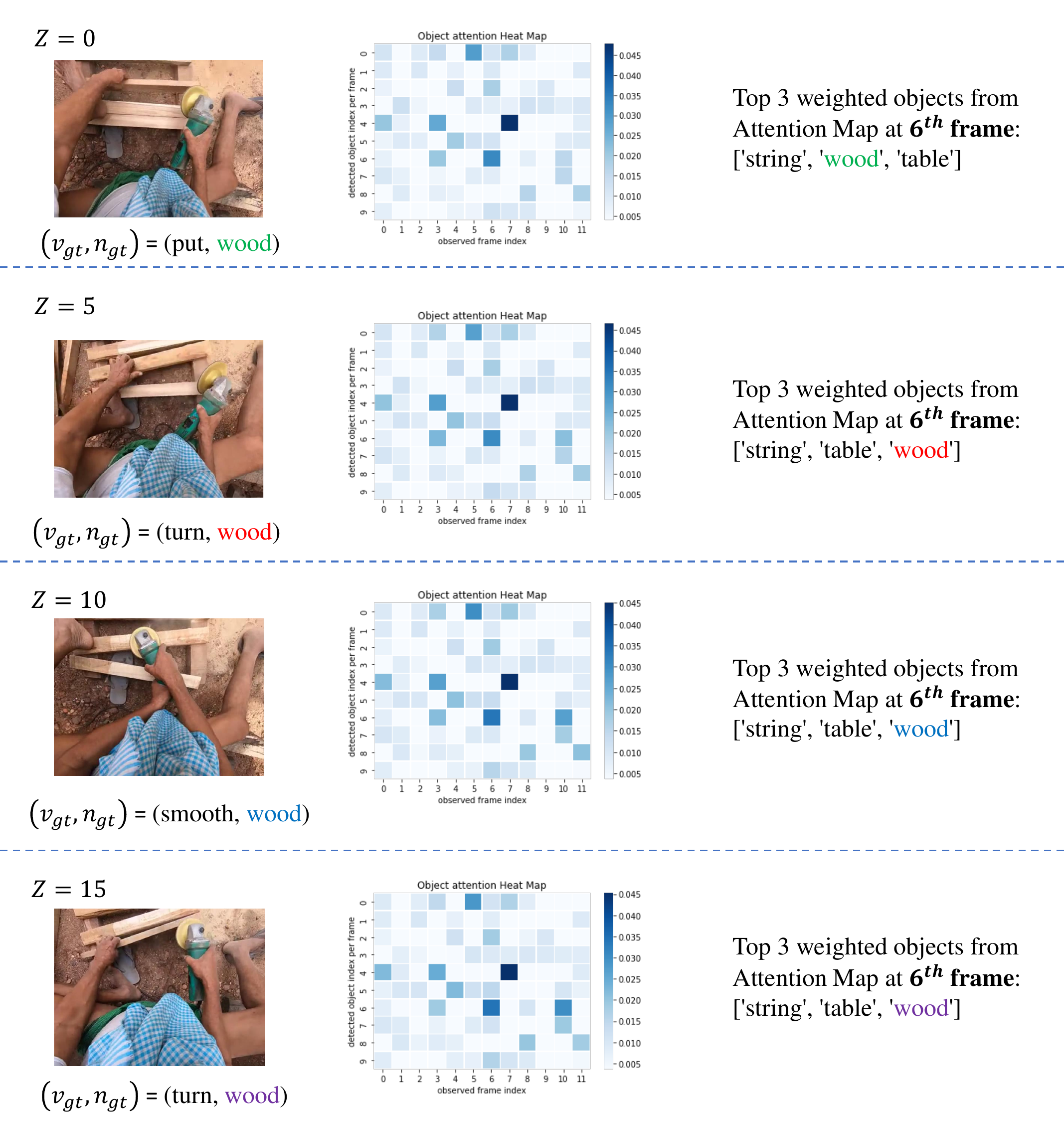}
  \end{subfigure}

  \caption{\textbf{Visualization of object attention heatmap and retrieved objects }, when $N_\text{seg}=3, N_\text{img}=4, N_\text{obj}=10, Z = 20$. \textbf{Left}: Representative video frame 
  correlate to $Z^{th}$ future step for anticipation, where $(v_{gt}, n_{gt})$ are ground truth action labels at $Z^{th}$ step in $(\text{verb}, \text{noun})$ pairs. \textbf{Middle}: Normalized object attention heatmap to previous observed visual input at $Z^{th}$ step. \textbf{Right}: Top 3 weighted objects from heatmap according to the center observed ($6^{th}$) frame.}
    \vspace{5em}
  \label{fig:heatmap6}
\end{figure*}

To better understand the impact of the object modality in the LTA task, we show qualitative examples of object attention weights produced by PTE for randomly selected validation examples in Figure~\ref{fig:heatmap1} to \ref{fig:heatmap6}. We use attention rollout~\cite{attentionrollout} to compute the overall attention weights from the output action predictions to the each object feature in the observed visual frames.

Specifically, we consider action predictions made at different future steps $Z=0$, $Z=5$, $Z=10$, and $Z=15$, and visualize their corresponding ``unrolled'' attention weights to the sequence of observed object features in 2D attention heatmaps (middle column in Figure~\ref{fig:heatmap1} to \ref{fig:heatmap6}). The x-axis of the 2D heatmap corresponds to the 12 observed visual frames, and the y-axis corresponds to the 10 objects retrieved with GLIP \cite{glip} for that frame. The object indices are randomly assigned, and the boxes corresponding to the same object may have different indices at each frame when they are visible. We observe that: First, our learned PTE models are often able to attend to the relevant objects in the observed frames, as the object categories of the top retrieved objects are often semantically similar to the ground truth objects for the target frames (we show the top 3 retrieved object categories for the center observed frame in the rightmost columns of Figure~\ref{fig:heatmap1} to \ref{fig:heatmap6}). Second, we observe that the attention heatmaps are relatively stable across different $Z$s, which is in line with our observation that there is a strong temporal continuity for the ground truth objects at different time steps. Overall, the visualizations confirm that object features are effectively used by our model as clues on highlighting possible interacted objects (nouns).

\begin{figure*}
  \centering
  \begin{subfigure}{\linewidth}
  \centering
    \includegraphics[width=\linewidth,page=2]{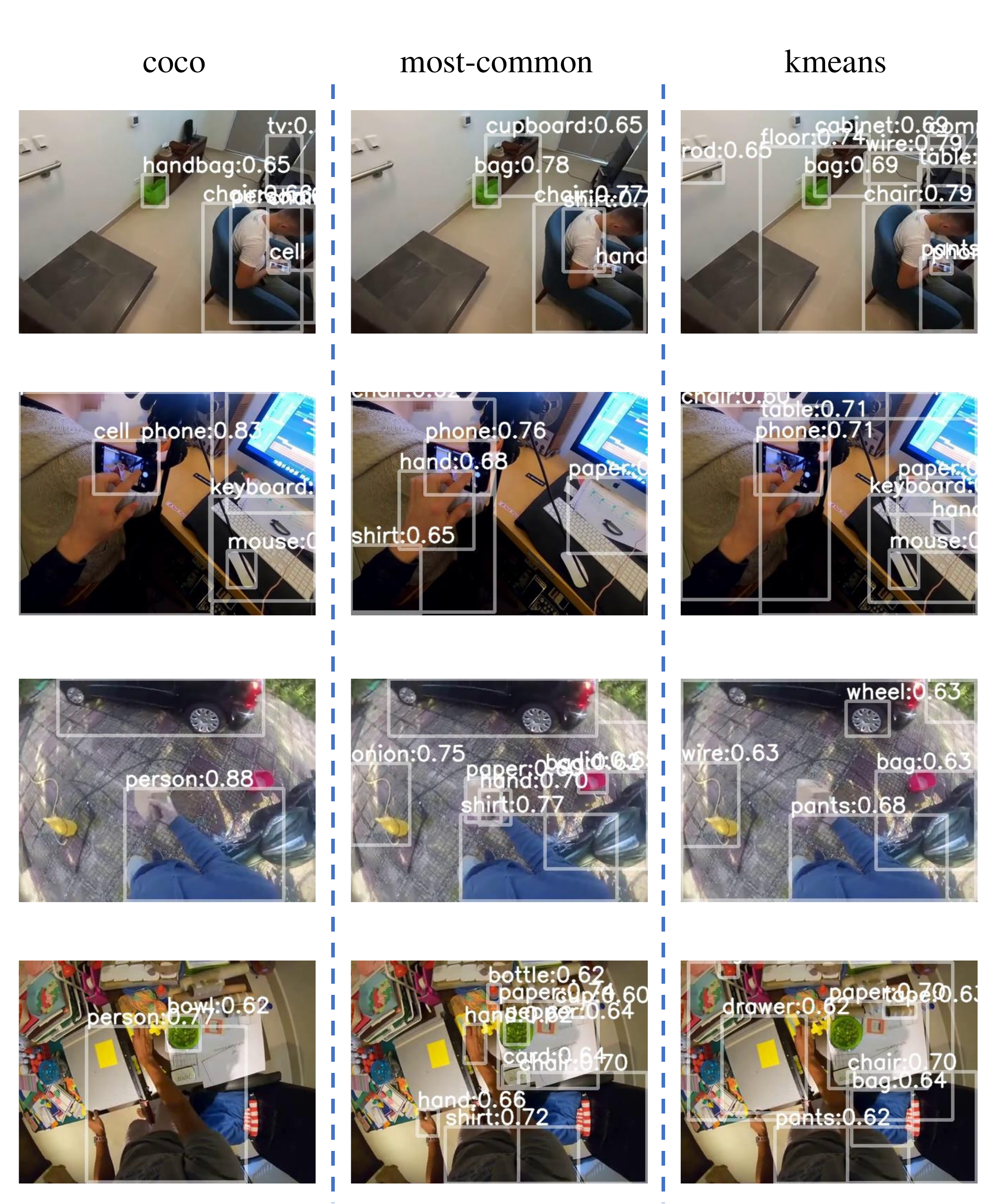}
  \end{subfigure}
  \vspace{-3mm}
  \caption{\textbf{Example GLIP detection results with different object prompts} on randomly sampled video frames. The K-Means clustering strategy (last column) offers the best precision and recall of relevant objects in all cases and offers the best generalization facing different scenarios.}
  \label{fig:prompt_visual1}
\end{figure*}

\begin{figure*}
  \centering
  \begin{subfigure}{\linewidth}
  \centering
    \includegraphics[width=\linewidth,page=3]{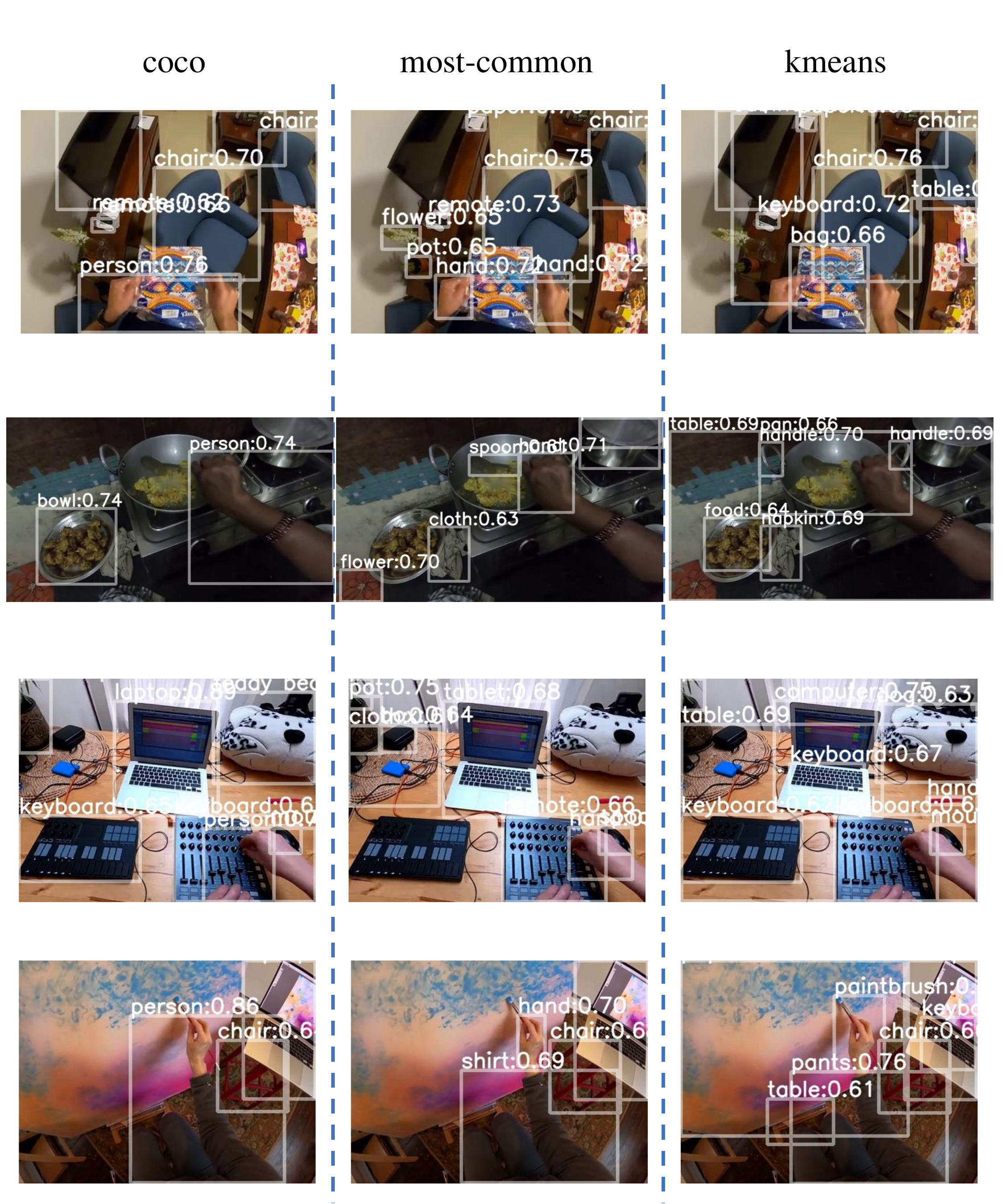}
  \end{subfigure}
  \vspace{-3mm}
  \caption{\textbf{Example GLIP detection results with different object prompts} on randomly sampled video frames. The K-Means clustering strategy (last column) offers the best precision and recall of relevant objects in all cases and offers the best generalization facing different scenarios.}
  \label{fig:prompt_visual2}
\end{figure*}

\begin{figure*}
  \centering
  \begin{subfigure}{\linewidth}
  \centering
    \includegraphics[width=\linewidth,page=4]{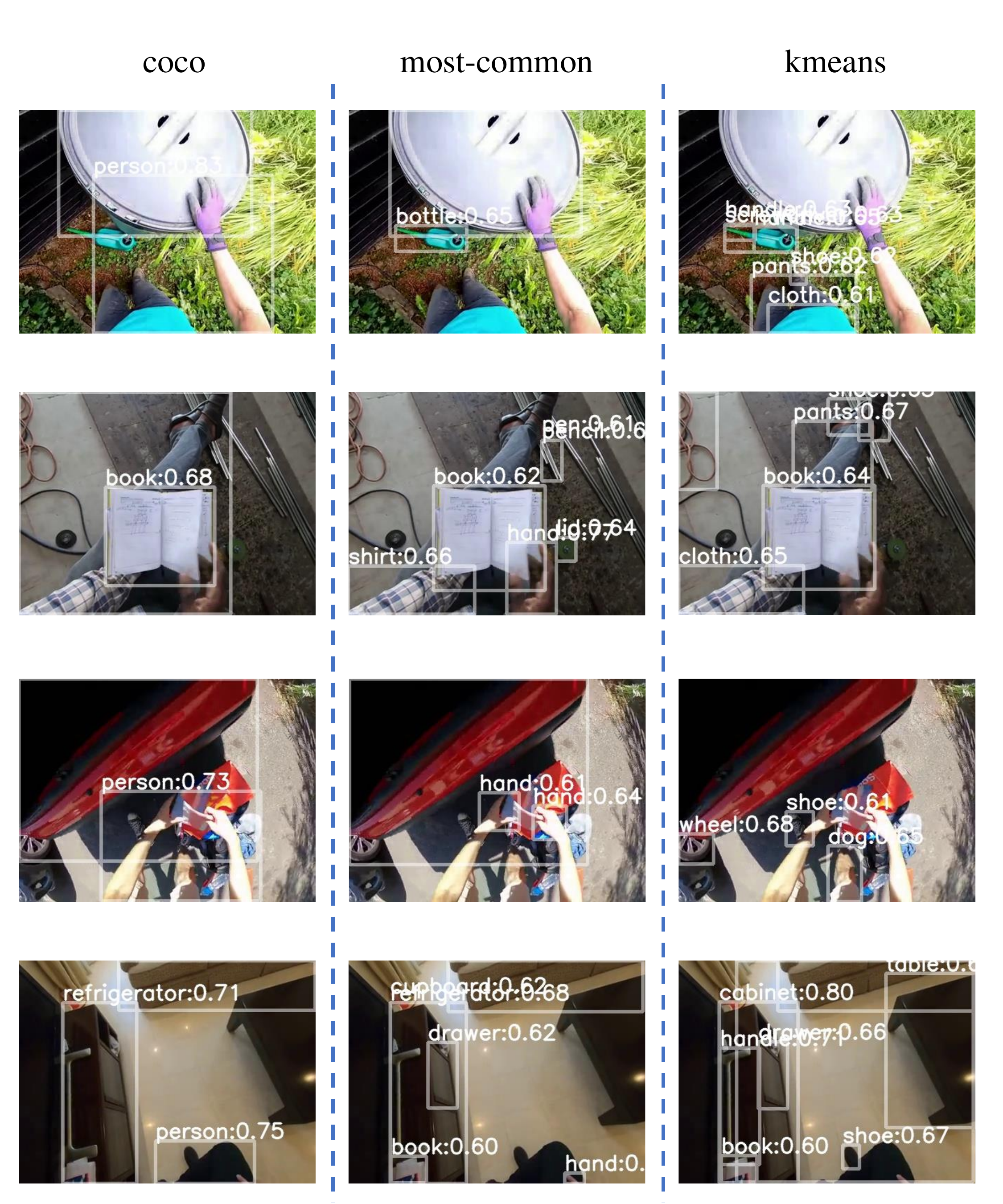}
  \end{subfigure}
  \vspace{-3mm}
  \caption{\textbf{Example GLIP detection results with different object prompts} on randomly sampled video frames. The K-Means clustering strategy (last column) offers the best precision and recall of relevant objects in all cases and offers the best generalization facing different scenarios.}
  \label{fig:prompt_visual3}
\end{figure*}

\begin{figure*}
  \centering
  \begin{subfigure}{\linewidth}
  \centering
    \includegraphics[width=\linewidth,page=5]{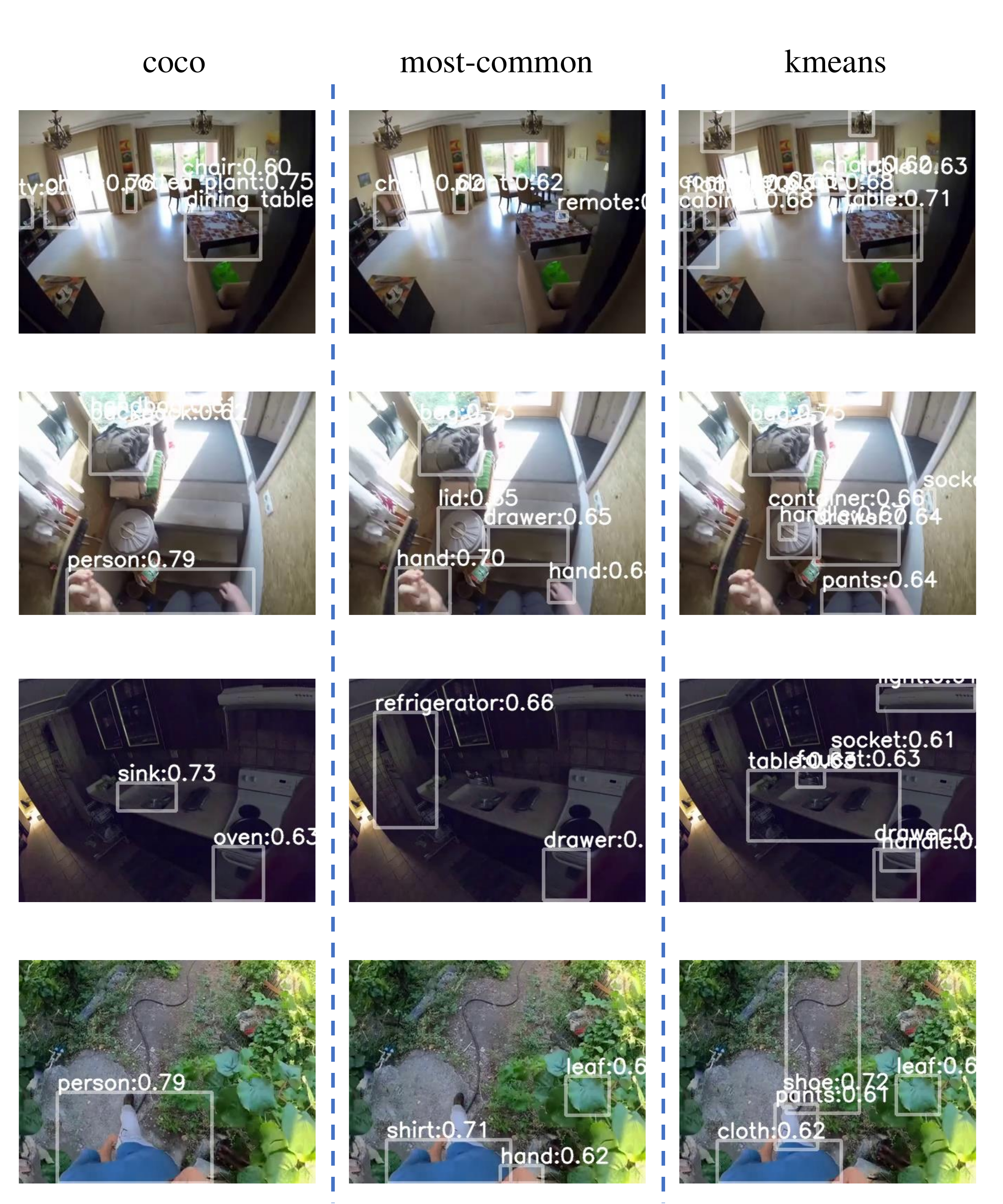}
  \end{subfigure}
  \vspace{-3mm}
  \caption{\textbf{Example GLIP detection results with different object prompts} on randomly sampled video frames. The K-Means clustering strategy (last column) offers the best precision and recall of relevant objects in all cases and offers the best generalization facing different scenarios.}
  \label{fig:prompt_visual4}
\end{figure*}

\subsection{Qualitative Analysis for Object Prompts}

Selecting optimal object prompts is critical for retrieving object features with GLIP \cite{glip}. We show more examples of the detection results with different objects on randomly sampled validation video frames. From Figure~\ref{fig:prompt_visual1} to \ref{fig:prompt_visual4}, we can observe that among the three, ``coco'' appears to be the least relevant and offer the least meaningful predictions across different scenarios, which is in line with our intuition and the empirical results. Compared to ``coco'', we can see that both ``most-common'' and ``kmeans'' give significantly more meaningful boxes than ``coco''. In addition, their predictions contain more important objects which are closely related to the LTA task. Overall, we observe that the detected objects with the ``kmeans'' object prompts provide the best precision-recall tradeoff for the sampled video frames among the three object prompt strategies. This is in line with the quantitative results when applying their corresponding detected objects to the LTA task.

\end{document}